# RENDITION:
# RECLAIMING WHAT A BLACK BOX TAKES AWAY

PEYMAN MILANFAR

GOOGLE RESEARCH

**Abstract.** The premise of our work is deceptively familiar: A black box $f(\cdot)$ has altered an image $\mathbf{x} \to f(\mathbf{x})$. Recover the image $\mathbf{x}$. This black box might be any number of simple or complicated things: a linear or non-linear filter, some app on your phone, etc. The latter is a good canonical example for the problem we address: Given only "the app" and an image produced *by* the app, find the image that was fed *to* the app. You can run the given image (or any other image) through the app as many times as you like, but you can not look inside the (code for the) app to see how it works

At first blush, the problem sounds a lot like a standard inverse problem [14, 17], but it is not in the following sense: While we have access to the black box $f(\cdot)$ and can run any image through it and observe the output, we do not know *how* the block box alters the image. Therefore we have no explicit form or model of $f(\cdot)$. Nor are we necessarily interested in the internal workings of the black box. We are simply happy to reverse its effect on a particular image, to whatver extent possible. This is what we call the "rendition" (rather than restoration) problem, as it does not fit the mold of an inverse problem (blind or otherwise).

We describe general conditions under which this *rendition* is possible, and provide a remarkably simple algorithm that works for both *contractive* and *expansive* black box operators. The principal and novel take-away message from our work is this surprising fact: One simple algorithm can reliably undo a wide class of (not too violent) image distortions.

**1. Introduction.** Let's consider an image[1] $\mathbf{x}^*$ that has been transformed to another image $f(\mathbf{x}^*)$ by an operator $f : [0,1]^N \to [0,1]^N$ that can be evaluated, but is otherwise not known, specified parametrically, or otherwise. The problem at hand is this: Given $f(\mathbf{x}^*)$, render an approximation to $\mathbf{x}^*$.

A general approach to this problem is brute force learning or high-dimensional regression: since we have free access to the black box $f(\cdot)$, we can push many images through it, collect the pairs $(\mathbf{x}_i, f(\mathbf{x}_i))$, and build a (possibly deep) model that is sophisiticated enough to generalize and render any input with high accuracy, given an output not seen before. This approach has been tried in [8], but for a very limited use case of restoring a portrait image from unknown global operations on human faces. They only tackled the two issues of skin smoothing and skin color change. We choose to be much more general, and through a series of approximations arrive at a simple solution that is very generally applicable.

A more obvious and direct approach to this problem would be to find the inverse function $f^{-1}(f(\mathbf{x}^*))$. This may look simple, but it is not easy for several reasons. First, we remind the reader that direct inversion is a luxury we do not enjoy since we do not have an explicit description of $f$. All we are allowed to do is to evaluate $f(\cdot)$ in the "forward" direction. Second, it is far from clear whether the inverse of such a black box even exists – especially, one that alters information or frequency content (e.g. blurring or sharpening).

In the standard literature on inverse problems, two additional conditions provide a framework that makes the problem more tractable: First, the "forward" operator $f(\mathbf{x})$ is explicitly specified (with perhaps some unknown parameters). It is generally a known physical model, or a filter with known structure (e.g. blind or non-blind blur, whether spatially varying or spatially invariant). And second, prior information in the form of a probability density $p(\mathbf{x})$ on the likely class of input images is assumed to be given. In our present treatment, we do not assume either of these conditions,

---

[1]Lexicographically scanned into a vector of length $N$

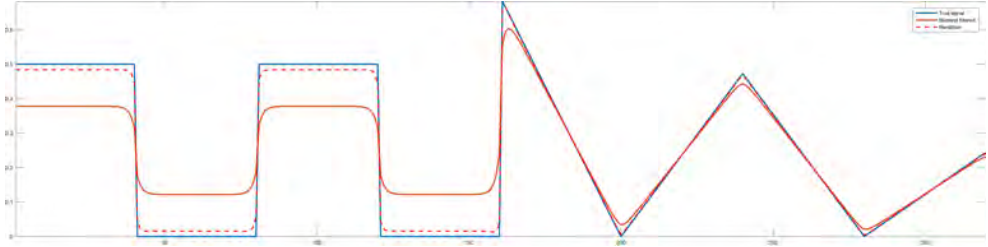

FIG. 1. *A one dimensional example of rendition: the blue curve is the underlying signal $\mathbf{x}^*$, and the red solid line is the distorted version $f(\mathbf{x}^*)$ computed by applying a bilateral filter to $\mathbf{x}^*$. The dashed red line illustrates the rendition.*

but their inclusion can be the subject of future investigations.

Instead, we will specify conditions under which we can render a solution $\widehat{\mathbf{x}} \approx \mathbf{x}^*$ with only access to "evaluation" of a black box operator $f(\mathbf{x})$. Examples of use cases that fit our approach in practice are common and numerous, though interestingly they have not been treated widely in the literature. One area of increasing importance is mobile photography and photo editing applications (i.e. apps). Photoshop and Lightroom have been around for a long time, but their use was largely limited to (semi-) professional photographers and enthusiasts who could afford expensive gear and post-processing software packages. Nowadays, many cheap or free mobile apps (indeed the camera apps themselves) enable users to edit images in non-trivial ways. Both the users themselves and the recipients of these processed images may want to revert at least some of these effect and obtain a rendition closer to the original. In some applications this functionality is enabled by saving multiple copies of the image (the original and the processed) so that the user can easily revert back. Rendition provides an approximate (but storage friendly) alternative.

Another possible area of interest is forensics. In security, law enforcement, and forensic investigations, the source hardware of the given visual material may be inaccessible directly, or hidden from the analyst. For instance, a blurry photo of a scene obtained from a remote security camera may be available, where neither the make nor the parameters of the camera are known. While it may be possible to render additional images from the sensor, and even have a software application that mimics the sensor, the operator is interested to render an approximation to the original scene without any specific knowledge of the camera.

To be more general, in the context of our approach, the idea is that given any accessible "forward" simulation of the process that leads to the altered image, we can render an approximation to the original image of the scene. To afford some intuition, we present a one-dimensional example of rendition shown in Fig. 1. In this example, the blue curve is the underlying signal $\mathbf{x}^*$, and the red solid line is the distorted version $f(\mathbf{x}^*)$ given by applying a non-linear edge-preserving(bilateral) filter to $\mathbf{x}^*$. The dashed red line illustrates the rendition, which amounts to undoing the effect of the bilateral filter.

**2. Formulation of A Loss for The Rendition Problem.** To *render* an input image given an output, we proceed by defining a loss function that takes advantage of correlations between the inputs $\mathbf{x}$, the outputs $f(\mathbf{x})$, and the respective residuals.

Let's define the (observable) residuals as follows:

$$\mathbf{r} = f(\mathbf{x}) - f(\mathbf{x}^*), \tag{2.1}$$

and consider the (cross-) correlation function

$$c(\mathbf{x}, \mathbf{x}^*) = \mathbf{x}^T \mathbf{r} = \mathbf{x}^T (f(\mathbf{x}) - f(\mathbf{x}^*)). \tag{2.2}$$

This loss function on its own is not powerful enough to give a reliable solution because there are at least three ways in which it could be driven to zero:

- If $\mathbf{x} = \mathbf{0}$. This is a degenerate and undesirable solution which we should like to avoid.
- If the residual is zero. This would imply that we are at a solution of the equation $f(\mathbf{x}) = f(\mathbf{x}^*)$, which under some regularity conditions may imply that $\mathbf{x}$ is near $\mathbf{x}^*$. And finally,
- If the residual is orthogonal to the input – a feature that implies that the residual behaves either like white noise, or it does not contain elements from the image itself. The concept of orthogonality of the residual to the signal is well-known. It has been harnessed successfully by many algorithms such as the Dantzig Selector [5], boosting techniques for denoising [6, 19, 22, 28, 13], etc. Indeed, enforcing orthogonality between the signal and output residual is the underlying force behind the normal equations in statistical estimation (e.g., least squares and Kalman filtering).

In order to avoid the useless trivial solution $\mathbf{x} = \mathbf{0}$, and encourage the other conditions, we formulate a second term that we shall attempt to *maximize*

$$a(\mathbf{x}) = \frac{1}{2} \mathbf{x}^T \mathbf{x}. \tag{2.3}$$

To summarize, we define the ***rendition loss*** as the composite function

$$\phi(\mathbf{x}) = c(\mathbf{x}, \mathbf{x}^*) - a(\mathbf{x}) = \mathbf{x}^T (f(\mathbf{x}) - f(\mathbf{x}^*)) - \frac{1}{2} \mathbf{x}^T \mathbf{x} \tag{2.4}$$

We will hence attempt to solve the following optimization problem:

$$\min_{\mathbf{x}} \phi(\mathbf{x}). \tag{2.5}$$

Next, we discuss conditions under which this loss yields a desired result, and give an iterative algorithm for solving the rendition problem. As a preview of coming attractions, we will show that quite surprisingly, the above optimization problem can be addressed with a very simple and convergent approximate gradient descent iteration as long as the operator $f(\cdot)$ is Lipschitz continuous (but not necessarily contractive). The iteration furthermore involves only the residuals as follows:

$$\mathbf{x}_{k+1} = \beta \mathbf{x}_k - \gamma \left[ f(\mathbf{x}_k) - f(\mathbf{x}^*) \right] \tag{2.6}$$

This solution is not obvious; its derivation will require some detailed analysis which we provide in the next sections.

**3. Solution of the Rendition Problem.** Let's for the moment assume that $f(\mathbf{x})$ is differentiable (an assumption we will relax shortly). The gradient of the proposed loss is:

$$\nabla \phi(\mathbf{x}) = \nabla (\mathbf{x}^T f(\mathbf{x})) - f(\mathbf{x}^*) - \mathbf{x} \tag{3.1}$$

$$= f(\mathbf{x}) + \nabla f(\mathbf{x})^T \cdot \mathbf{x} - f(\mathbf{x}^*) - \mathbf{x} \tag{3.2}$$

**Remark:** The gradient $\nabla f(\mathbf{x})$ will play a starring role in the rest of the paper, so it is important to discuss its properties in detail here. Most operators $f(\mathbf{x})$ of interest are a combination of operations that are either pointwise, such as tone-mappers or color transformations; or pseudo-linear, such as the bilateral or non-local means; or the result of minimizing a particular (implicit or explicit) cost function. If the operator $f(\mathbf{x})$ acts pointwise on $\mathbf{x}$, then the gradient $\nabla f(\mathbf{x})$ is a diagonal matrix (and therefore also symmetric.) When the operator is implemented in pseudo-linear form $f(\mathbf{x}) = \mathbf{W}(\mathbf{x})\mathbf{x}$, the weights matrix $\mathbf{W}$ can be closely approximated by a symmetric matrix [18, 20]; and in this case [23], $\nabla f(\mathbf{x}) = \mathbf{W}(\mathbf{x})$. More generally, whenever the operator can be written as the gradient of some scalar loss [26, 19], say $f(\mathbf{x}) = \nabla \Theta(\mathbf{x})$, it describes a conservative (or curl-free) vector field. This, in turn, means that the gradient $\nabla f(\mathbf{x}) = \nabla^2 \Theta(\mathbf{x})$ of the operator, being the *Hessian* of $\Theta(\mathbf{x})$, is by definition a symmetric matrix. Most denoisers, regularizers, and filters built upon them fall into this category [19, 23]; as do proximal operators. The latter case is instructive:

$$(3.3) \qquad f(\mathbf{x}) = \arg\min_{\mathbf{u}} \frac{1}{2}\|\mathbf{u} - \mathbf{x}\|^2 + \alpha\,\theta(\mathbf{u}).$$

We can compute the gradient of the right-hand side, solve symbolically for $f(\mathbf{x})$, and approximate for small $\alpha$ (i.e. consistent with our assumption of "gentle" operators):

$$(3.4) \qquad f(\mathbf{x}) = (\mathbf{I} + \alpha\,\nabla\theta)^{-1}(\mathbf{x})$$
$$(3.5) \qquad \approx \mathbf{x} - \alpha\,\nabla\theta(\mathbf{x})$$
$$(3.6) \qquad = \nabla\left[\frac{1}{2}\mathbf{x}^T\mathbf{x} - \alpha\theta(\mathbf{x})\right].$$

We see that $f(\mathbf{x})$ is the gradient of the (scalar) term inside the brackets, which means that $\nabla f(\mathbf{x})$ is a symmetric matrix [2].

To summarize, in the remarks above we've argued that for all practical purposes, we can assume that $\nabla f(\mathbf{x})^T = \nabla f(\mathbf{x})$. With this in hand, we write

$$(3.7) \qquad \nabla\phi(\mathbf{x}) = f(\mathbf{x}) + \nabla f(\mathbf{x}) \cdot \mathbf{x} - f(\mathbf{x}^*) - \mathbf{x}$$

Next, at least in theory, and without (yet) any guarantee of convergence, we can consider a gradient descent algorithm based on this expression as follows:

$$(3.8) \qquad \mathbf{x}_{k+1} = \mathbf{x}_k - \gamma\nabla\phi(\mathbf{x}_k).$$

The right-hand-side still includes the term $\nabla f(\mathbf{x}) \cdot \mathbf{x}$, which contains the (inaccessible) gradient of $f(\mathbf{x})$. This would seem like a dead end. However, while we do not have access to the gradient of $f(\mathbf{x})$ directly, we can approximate the *directional derivative* as follows [23]:

$$(3.9) \qquad \nabla f(\mathbf{x}) \cdot \mathbf{d} \approx \frac{f(\mathbf{x} + \epsilon\mathbf{d}) - f(\mathbf{x})}{\epsilon},$$

for sufficiently small $\epsilon$. In particular we are interested in the case where $\mathbf{d} = \mathbf{x}$, which is computable with two activations of the function $f(\cdot)$ as follows:

$$(3.10) \qquad \nabla f(\mathbf{x}) \cdot \mathbf{x} \approx \frac{f(\mathbf{x} + \epsilon\mathbf{x}) - f(\mathbf{x})}{\epsilon} = \frac{f((1+\epsilon)\mathbf{x}) - f(\mathbf{x})}{\epsilon}.$$

---

[2]Even without resorting to approximation, such operators can be written exactly as $f(\mathbf{x}) = \mathbf{x} - \alpha\nabla\phi_s(\mathbf{x})$, where $\phi_s$ is a smoothed version of $\phi$. More specifically, $\phi_s$ is the Moreau envelope [21] of $\phi$.

From a practical standpoint, we can still proceed since all the quantities in the gradient descent equation can be computed, albeit inefficiently. As we will show later, we can simplify the computations even further to avoid two activations of $f(\cdot)$. For now, the more critical question is whether this iteration converge at all. And if so, to what? To address the quesiton of point of convergence, let's examine where the gradient vanishes or becomes small at a point near $\widehat{\mathbf{x}}$:

$$\nabla \phi(\widehat{\mathbf{x}}) = f(\widehat{\mathbf{x}}) + \nabla f(\widehat{\mathbf{x}}) \cdot \widehat{\mathbf{x}} - f(\mathbf{x}^*) - \widehat{\mathbf{x}} = 0 \tag{3.11}$$

$$\implies f(\widehat{\mathbf{x}}) - f(\mathbf{x}^*) = \nabla f(\widehat{\mathbf{x}}) \cdot \widehat{\mathbf{x}} - \widehat{\mathbf{x}} \tag{3.12}$$

$$\implies f(\widehat{\mathbf{x}}) - f(\mathbf{x}^*) = (\nabla f(\widehat{\mathbf{x}}) - \mathbf{I}) \widehat{\mathbf{x}} \tag{3.13}$$

Let's look at the spectrum of the operator on the right-hand side. Since $\nabla f(\widehat{\mathbf{x}})$ is assumed symmetric, it has real eigenvalues $\lambda_i$ and spectral radius $M = \max_i |\lambda_i|$. The eigenvalues of $(\nabla f(\widehat{\mathbf{x}}) - \mathbf{I})$ are $\lambda_i - 1$ and its spectral radius is $M' = \max_i |\lambda_i - 1|$. This can help us see how far from the solution we are likely to be:

$$\|f(\widehat{\mathbf{x}}) - f(\mathbf{x}^*)\| = \|(\nabla f(\widehat{\mathbf{x}}) - \mathbf{I}) \widehat{\mathbf{x}}\| \tag{3.14}$$

$$\leq \|(\nabla f(\widehat{\mathbf{x}}) - \mathbf{I})\| \|\widehat{\mathbf{x}}\| \tag{3.15}$$

$$\leq M' \|\widehat{\mathbf{x}}\|, \tag{3.16}$$

where $|M - 1| \leq M' \leq M + 1$. When the spectral norm $M$ is near 1, the proposed iteration will not drift far from the solution of the fixed point problem[3]. As $M$ gets much larger than 1, it is less likely that the solution $\widehat{\mathbf{x}}$ will be near the desired solution $\mathbf{x}^*$. Nevertheless, if $M$ is sufficiently close to 1, we have reason to be optimistic that we will obtain a useful rendition. Put more informally, one can imagine that if the effect of $f(\cdot)$ is severe, we are unlikely to render a good solution without additional constraints or information. However, if the function is reasonably well-behaved in the vicinity of $\mathbf{x}^*$ (in the sense of $M \approx 1$), we may succeed.

Given $f(\widehat{\mathbf{x}})$, the ability to render a solution that is useful in the sense that $\widehat{\mathbf{x}} \approx \mathbf{x}^*$ hinges, unsurprisingly, on the properties of $f$. First, and at a basic level, we assume that the black box $f(\mathbf{x})$ does not alter a null image. That is to say, we assume $f(\mathbf{0}) = \mathbf{0}$. Second, we mentioned above the gradient of $f(\mathbf{x})$ having bounded spectral radius. Since the gradient may not strictly exist everywhere for all filters, this notion will instead be invoked in weaker form below. What we are describing here is a kind of local "regularity" that can be formalized using the concept of Lipschitz continuity. This motivates the following assumption on the black box $f(\cdot)$:

**Assumption 1: $f(\mathbf{x})$ is Locally Lipschitz:** This means that for any $\mathbf{x}$ and $\mathbf{x}'$ in a small neighborhood $\mathcal{N}(\mathbf{x}^*)$ of $\mathbf{x}^*$, the operator $f(\mathbf{x})$ satisfies

$$\|f(\mathbf{x}) - f(\mathbf{x}')\| \leq M \|\mathbf{x} - \mathbf{x}'\|, \tag{3.17}$$

where $M$ is a positive constant[4].

A Lipschitz function is limited in how fast it can change locally (as dictated by $M$). While Lipschitz property does not imply differentiability, it does provide (through Rademacher's Theorem [11]), a guarantee that the function will be almost everywhere (a.e.) differentiable; that is, if not at $\mathbf{x}^*$, then near it. When the function

---

[3] As shown in [23], when $f(\mathbf{x})$ is a smoothing filter, its gradient $\nabla f(\mathbf{x})$ has spectral radius $M \leq 1$.
[4] The choice of norm is somewhat unimportant, but for our purposes we will use the $\ell_2$ norm throughout the paper.

*is* differentiable, $M$ provides a bound on the derivative. More intuitively, the Lipschitz property guarantees that the effect of the operator is not strong (again as dictated by a sufficiently small $M$) – that it does not change the input image too severely. As a result of this guarantee, as we shall see below, we are able to render an approximation to the unseen input given only the output and activations of the function $f(\cdot)$.

**Operators That Satisfy Assumption 1:** The approach we propose here would not be very useful if it could only be applied to a small number of operators. But quite the contrary is true. In fact, nearly every operation on images one can conjure up satisfies the Lipschitz property (and often with a constant $M$ near 1).

- *Smoothing/Blurring:* To begin, we note that in [23] it was shown that a vast array of (linear and nonlinear) smoothing filters commonly used in the literature are *non-expansive*. That is, they are Lipschitz with $M \leq 1$. These include linear denoisers with Tikhonov regularization [10], and Wiener filtering [10]; sparsity-based methods such as K-SVD [1], patch based methods such as BM3D [9], the domain transform filter [12], the Bilater filter [29, 24], and Non-local Means (NLM) [26]; Gaussian mixture models such as EPLL [31], TNRD [7].
- *Sharpening/Deconvolution:* As we've described in [23, 19] and elsewhere, the contractive operators are the building blocks of many other more sophisticated (non-contractive) operators that are used for various enhancement purposes including sharpening, deconvolution, super-resolution, etc. Consider a smoothing operator $f(\mathbf{x})$ (say the bilateral filter). Analogous to the linear unsharp mask, we can construct a spatially adaptive sharpening filter as $g(\mathbf{x}) = \mathbf{x} + \alpha(\mathbf{x} - f(\mathbf{x}))$. Sum of two Lipschitz functions with constants $M_1$ and $M_2$ is also Lipschitz with corresponding constant $M_1 + M_2$. Furthermore, the Lipschitz constant of the function $cf(\mathbf{x})$ is $|c|M$. Therefore the "sharpening" operator $g(\mathbf{x})$ has Lipschitz constant $M_g \leq (1 + 2|\alpha|)$, which may exceed unity, but not by much since we typically choose a small $\alpha$.
- *Composite Filters:* Generalizing the last group of filters, as described in [26, 27] and elsewhere, we can construct a variety of non-linear filters with a desirable range of effects in several ways. The simplest is to use a polynomial application of a base filter[5] $f(\mathbf{x})$. To be more precise, if we describe $f(\mathbf{x}) = \mathbf{W}(\mathbf{x})\mathbf{x}$, then we can generate $g(\mathbf{x}) = p_n(\mathbf{W}(\mathbf{x}))\mathbf{x}$, where $p_n(\lambda) = \sum_{j=0}^{n} \alpha_j \lambda^j$ is a polynomial of degree $n$. If the base filter is 1-Lipschitz, then the filter $g(\mathbf{x})$ has Lipschitz constant $M_g \leq \sum_{j=0}^{n} |\alpha_j|$, which one could control by keeping the $\alpha_i$'s small in magnitude.
- *Tone-mapping and Gamma Correction:* Tone mapping functions are pixelwise operators that typically expand or compress the dynamic range of images for better viewing. For instance, a typical gamma curve[6] acts pointwise on the elements of $\mathbf{x}$ as $f(x_i) = x_i^\gamma$. More general tone-mapping curve are S-shaped curves that map $[0, 1]$ to $[0, 1]$. See for example the curves in Fig. 17. All these functions are Lipschitz, with increasing constants that depend on the sharpness of the transition at 0.5.
- *Compression and Compressed Sensing:* Classical tranform coding techniques (such as the jpg standard) apply an orthogonal transform (e.g. DCT) to the

---

[5]More generally, we can use the composition of any two Lipschitz functions $f_1(f_2)(\mathbf{x})$ each with constants $M_1, M_2$. This composition is another Lipschitz function with constant $M_1 M_2$

[6]For $\gamma < 1$ This function is only uniformly continouous, but is Lipschitz if we exclude the point $x = 0$

image (a 1-Lipschitz operator), then shrink the resulting coefficients (another 1-Lipschitz operator), and finally quantize these coefficients (an $M$-Lipschitz operator, where $M$ is inversely proportional to the number of quantization levels). Overall therefore, compression/decompression is an $M$-Lipschitz operator with $M$ possibly close to 1. In the compressed sensing literature, the operators that are considered are random matrices that satisfy a *restricted isometry* property [4] $(1-\delta)\|\mathbf{x}\|_2^2 \leq \|\Phi\mathbf{x}\|_2^2 \leq (1+\delta)\|\mathbf{x}\|_2^2$. This condition effectively guarantees with high probability [15] that the Lipschitz constant of the operator is inside the range $1 \pm 4\delta$

- *Neural Networks:* Each layer of a convolutional neural network is composed of a linear smoothing filter (a 1-Lipschitz operator) followed by a nonlinearity. If the nonlinearity is the classical sigmoid, this is of course 1-Lipschitz. If not, the ReLU nonlinearity over a compact range of values of the image ($[0, 1]$) still qualifies as a 1-Lipschitz operator. Between layers, various pooling operators can be employed. The overall effect of these and the corresponding effect on the Lipschitz constant of the overall network has been carefully studied recently in [3].

**Estimates of the Lipschitz Constant:** To estimate the Lipschitz constant for a given filter we can resort to approximations. First, let's recall that

$$(3.18) \qquad M = \max_{\mathbf{x},\mathbf{x}'\in[0,1]^N} \frac{\|f(\mathbf{x}) - f(\mathbf{x}')\|}{\|\mathbf{x} - \mathbf{x}'\|}$$

In order to estimate this quantity locally around some $\mathbf{x}$, we use the following empirical estimate

$$(3.19) \qquad \widehat{M} = \max_{\mathbf{x},\,\mathbf{d}} \frac{\|f(\mathbf{x}+\mathbf{d}) - f(\mathbf{x})\|}{\|\mathbf{d}\|}$$

where $\mathbf{x}$ and $\mathbf{d}$ are independent random vectors with uniform density $\mathbf{U}[0,1]^N$, and the maximum is computed over a large number of samples $(\mathbf{x}_i, \mathbf{d}_i)$. For now, consider a concrete example where an image is adaptively sharpened through a bilateral filter, and gamma-corrected as follows:

$$(3.20) \qquad f(\mathbf{x}) = (\mathbf{x} + \alpha(\mathbf{x} - \text{bilat}(\mathbf{x}, \sigma_1, \sigma_2))^\gamma$$

With $\alpha = 1$, $[\sigma_1, \sigma_2] = [10, 3]$, and $\gamma = 0.65$, we estimate $\widehat{M} = 1.027$. This sequence of nonlinear operations are shown in Fig. 2 for an example image. In Fig. 12, we will illustrate the rendition of the original image (left) from the altered one on the right.

In the appendix, we present more examples of these estimates for several operators. We verify that operators that are predominantly smoothers should yields values of $M$ smaller than 1, whereas those that have a sharpening effect should generally yield values larger than 1. The intuition to be gleaned from these estimates and how well the proposed algorithm is expected to work is as follows: The farther away the Lipschitz constant is from 1 (in either direction), the less likely it is for us to succeed in rendering a useful result. When $M$ is much smaller than 1, the black box operator's effect is too severe as a smoother and therefore the rendition will fail (imagine strong blurring for instance). On the other hand, if $M$ is much bigger than one, the black box has either added too much spurious information to the image or overwhelmed parts of the image (e.g. with saturation in an HDR scene) that rendition may again fail. These remarks are consistent with calculations of Lipschitz constants for various operators shown in the appendix.

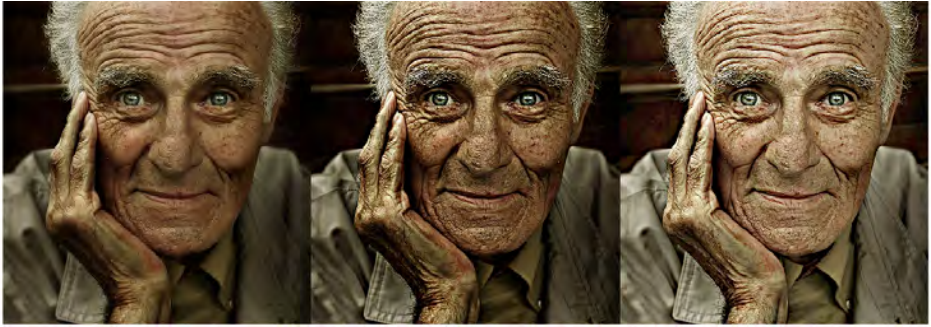

FIG. 2. *Left to right: Input image, Sharpened with Domain (Bilateral) Transform, tone-mapped by $\gamma$ correction. Can we go back? The answer is yes. In Fig. 12, we illustrate the rendition of the original image (left) from the altered one on the right. Credit: Andrzej Dragan*

**Assumption 2:** Earlier when we discussed the directional derivative of $f(\mathbf{x})$ and its approximation, we did not discuss pathologies where these quantities may not even exist. Here we will assume that $f(\mathbf{x})$ **has a non-trivial directional derivative at $\mathbf{x}^*$**. This assumption is not very restrictive in light of the broader definition of the directional derivative [16] we will give next. First recall that the directional derivative of a differentiable function in the direction $\mathbf{d}$ is defined as follows:

$$(3.21) \qquad \nabla f(\mathbf{x}) \cdot \mathbf{d} = \lim_{\epsilon \to 0} \frac{f(\mathbf{x} + \epsilon \mathbf{d}) - f(\mathbf{x})}{\epsilon}$$

When $f$ is only Lipschitz, a looser definition of the directional derivative applies. Following [16], we think of the directional derivative as a *set* $\nabla f(\mathbf{x}, d)$ which consists of all points $\mathbf{z}$ that are the limit points of the sequence

$$(3.22) \qquad \mathbf{z}_k = \frac{f(\mathbf{x}_k + \epsilon_k \mathbf{d}) - f(\mathbf{x}_k)}{\epsilon_k}$$

where $\mathbf{x}_k \to \mathbf{x}$ and $\epsilon_k \to 0$. To be explicit, this assumption ensures that the approximation to the directional derivative is well-defined in the neighborhood of $\mathbf{x}^*$.

In the scenario where the function is differentiable, this is equivalent to saying that the Jacobian is full rank and therefore invertible. And when this is true, the Inverse Function Theorem [2] guarantee that the function is locally 1-1 and therefore locally invertible. If instead there is a null-space to the Jacobian, all this means is that we cannot restore *some portions* of the original[7] $\mathbf{x}^*$. More to the point, in [16], a generalization of the Inverse Function Theorem guarantees the function is locally invertible when it is only Lipschitz (as in our case).

**Operators That Satisfy Assumption 2:** Let's consider the simplest case of linear filters $f(\mathbf{x}) = \mathbf{W} \mathbf{x}$. If $\mathbf{W}$ is a discretization of a simple blur, then depending on the structure of the blurring kernel and the chosen boundary condition, this matrix can certainly be rank deficient, but the dimension of the null-space would be small. For instance, if the blur kernel is a disk, which entirely kills certain spatial frequencies, then **W** would be rank deficient against components of the image at those spatial frequencies. In general, one would not be able to recover such frequencies unless strong prior information was available. In other cases, for instance Gaussian blur, the matrix is generally not rank deficient, but will become increasingly more ill-conditioned with wider kernels. So generically, it is hard to construct linear filters that completely annihilate typical images that contain a wide range of spatial frequencies. At worst, some small number, or range of spatial frequencies are knocked out and are not directly recoverable by rendition (i.e. a small null-space exists).

---

[7]In such a scenario, regularization [23] can be applied to still obtain a solution. We briefly discuss this notion is Section 5.

The situation for non-linear filters is analogous. Let's consider the case of (pseudo-linear) filters of the form $f(\mathbf{x}) = \mathbf{W}(\mathbf{x})\mathbf{x}$. As shown in [23], under some mild regularity conditions, $\nabla f(\mathbf{x}) = \mathbf{W}(\mathbf{x})$. The matrix $\mathbf{W}(\mathbf{x})$ may be constructed from any number of non-linear kernels such as bilateral or non-local means. As discussed at length in [26, 25], by looking at the eigendecomposition of the filter matrix for natural images, we can see that these nonlinear filters are generally full rank when the support of the filters is small (i.e. when the filters are *local*). When the filters are non-local (or even truly global), the numerical rank of the filter matrix drops, similarly to the linear case. So again, most locally operating nonlinear filters also do not annihilate most images completely. And non-local filters will annihilate only a subset of features in any image. So consistent with the main message of the paper, as long as the effect of the black box is not severe, we may stand a chance of obtaining a useful rendition. With the two assumptions above in place, we now have the key result that enables us to move forward.

THEOREM 3.1 ([16]). *The function $f(\mathbf{x})$ is locally Lipschitz invertible at $\mathbf{x}^*$ if and only if the above two conditions are satisfied at $\mathbf{x}^*$.*

The implications of this result are deep. It will enable us to invert a very large class of degradations as long as their effect is not too severe. A particularly noteworthy implication is that the local inverse is itself guaranteed to be locally Lipschitz (with a different positive constant $m$). That is, the function is locally *bi*-Lipschitz near $\mathbf{x}^*$:

$$(3.23) \qquad m \left\| \mathbf{x} - \mathbf{x}' \right\| \leq \left\| f(\mathbf{x}) - f(\mathbf{x}') \right\| \leq M \left\| \mathbf{x} - \mathbf{x}' \right\|.$$

Now let's return to the iterative algorithm we proposed earlier:

$$(3.24) \qquad \mathbf{x}_{k+1} = \mathbf{x}_k - \gamma \nabla \phi(\mathbf{x}_k),$$

where

$$(3.25) \quad \nabla \phi(\mathbf{x}) = f(\mathbf{x}) - f(\mathbf{x}^*) + \nabla f(\mathbf{x})\mathbf{x} - \mathbf{x} = f(\mathbf{x}) - f(\mathbf{x}^*) + (\nabla f(\mathbf{x}) - \mathbf{I})\mathbf{x}$$

As we noted before, the directional derivative $\nabla f(\mathbf{x})\mathbf{x}$ could be approximated with two activations of the filter $f(\mathbf{x})$. We wish to simplify even further to avoid this. Therefore, we instead appeal to an approximation of the gradient of the loss directly. Namely, we recall that $\left\| (\nabla f(\mathbf{x}) - \mathbf{I})\mathbf{x} \right\| \leq M' \|\mathbf{x}\|$. Therefore, as $M$ is near 1, we will approximate

$$(3.26) \qquad (\nabla f(\mathbf{x}) - \mathbf{I})\mathbf{x} \approx \mu \mathbf{x}$$

where $|M - 1| \leq \mu \leq M + 1$ is a non-negative scalar.

Now, recalling the definition of the loss function and its gradient:

$$(3.27) \qquad \phi(\mathbf{x}) = \mathbf{x}^T (f(\mathbf{x}) - f(\mathbf{x}^*)) - \frac{1}{2}\mathbf{x}^T \mathbf{x},$$

$$(3.28) \qquad \nabla \phi(\mathbf{x}) = f(\mathbf{x}) - f(\mathbf{x}^*) + (\nabla f(\mathbf{x}) - \mathbf{I})\mathbf{x},$$

we have the approximate gradient of the loss:

$$\widetilde{\nabla}\phi(\mathbf{x}) = f(\mathbf{x}) - f(\mathbf{x}^*) + \mu\,\mathbf{x} \tag{3.29}$$

This gives an elegantly simple approximate gradient descent iteration:

$$\mathbf{x}_{k+1} = \mathbf{x}_k - \gamma\widetilde{\nabla}\phi(\mathbf{x}_k) \tag{3.30}$$

$$\implies \mathbf{x}_{k+1} = (1-\gamma\mu)\mathbf{x}_k - \gamma\left[f(\mathbf{x}_k) - f(\mathbf{x}^*)\right] \tag{3.31}$$

With a sufficiently small step size $\gamma$, the factor $(1 - \gamma\mu)$ is slightly less than one, and has a mild dampening effect. This iteration overall has a nice intuitive interpretation: dampen the last estimate slightly, and move repeatedly in the direction of the residual. This interpretation is reminiscent of (but distinct from) the concept of *boosting* described in several recent works [23, 22, 19, 6, 28]. We should note that with the exception of the first term on the right-hand side, the equation (3.31) is a damped fixed point iteration for solving the equation $f(\mathbf{x}_k) = f(\mathbf{x}^*)$. We hasten to note that such a fixed point iteration would *only* converge for contractive operators. The interesting property of the proposed approach is that this simple iteration works toward minimizing $\phi(\mathbf{x})$ even if the operator $f(\mathbf{x})$ is expansive. We discuss the issue of convergence next.

**3.1. Convergence.** Is the iteration (3.31) convergent? In order to answer this question, we need to examine when the operator $\psi(\mathbf{x}) = \mathbf{x} - \gamma\widetilde{\nabla}\phi(\mathbf{x})$ on the right-hand side is contractive. We compute its gradient:

$$\nabla\psi(\mathbf{x}) = (1-\gamma\mu)\mathbf{I} - \gamma\nabla f(\mathbf{x}) \tag{3.32}$$

If the spectral radius of $\nabla\psi(\mathbf{x})$ is below 1, we have convergence. This spectal radius is bounded above as:

$$\rho(\nabla\psi(\mathbf{x})) = |1 - \gamma\mu| + \gamma M \tag{3.33}$$

which must satisfy

$$|1 - \gamma\mu| + \gamma M < 1 \implies \gamma < \frac{2}{\mu + M} \tag{3.34}$$

This is a sufficient condition to ensure the convergence of the proposed iteration to a local optimum.

**Remark**: While the loss function $\phi(\mathbf{x})$ is not convex, it is interesting to note that it is *locally, directionally* convex. We can look at a Taylor series expansion around the optimum $\widehat{\mathbf{x}}$, where $\nabla\phi(\widehat{\mathbf{x}}) = 0$. Let's consider a small perturbation $\widehat{\mathbf{x}} \to \widehat{\mathbf{x}} + \epsilon\mathbf{d}$ in an arbitrary direction $\mathbf{d}$. For sufficiently small $\epsilon$, we have

$$\phi(\widehat{\mathbf{x}} + \epsilon\mathbf{d}) = \phi(\widehat{\mathbf{x}}) + \epsilon\,\nabla\phi(\widehat{\mathbf{x}})\mathbf{d} + \frac{\epsilon^2}{2}\,\mathbf{d}^T\nabla^2\phi(\widehat{\mathbf{x}})\mathbf{d} \tag{3.35}$$

$$= \phi(\widehat{\mathbf{x}}) + \frac{\epsilon^2}{2}\,\mathbf{d}^T\nabla^2\phi(\widehat{\mathbf{x}})\mathbf{d} \tag{3.36}$$

$$\approx \phi(\widehat{\mathbf{x}}) + \frac{\epsilon^2}{2}\,\mathbf{d}^T\widetilde{\nabla}^2\phi(\widehat{\mathbf{x}})\mathbf{d} \tag{3.37}$$

$$= \phi(\widehat{\mathbf{x}}) + \frac{\epsilon^2}{2}\,\mathbf{d}^T(\nabla f(\widehat{\mathbf{x}}) + \mu\mathbf{I})\mathbf{d} \tag{3.38}$$

$$= \phi(\widehat{\mathbf{x}}) + \frac{\mu\epsilon^2}{2}\,\mathbf{d}^T\mathbf{d} + \frac{\epsilon^2}{2}\,\mathbf{d}^T\nabla f(\widehat{\mathbf{x}})\mathbf{d} \tag{3.39}$$

We observe that as we move away from $\mathbf{x}^*$ in an arbitrary direction $\mathbf{d}$, the loss may increase or decrese depending directly on whether the gradient $\nabla f(\widehat{\mathbf{x}})$ is positive definite (and therefore the quadratic form positive or negative valued) in that direction. When the filter $f(\mathbf{x})$ is strictly of denoising/smoothing type, all eigenvalues of the filter are non-negative [19, 23], and we can be assured that $\nabla f(\mathbf{x})$ is non-negative definite in any direction. However, for more general filters we've considered here in the context of rendition, this is not necessarily the case. In such instances, the convergence may be to a saddle point. When the perturbation is small, this solution is often enough to give a good rendition, and we will stop the iterations. A reliable approach to remain in this useful local critical point is to pick a small step size $\gamma$, and iterate roughly $k = \frac{1}{\gamma M}$ times. Alternatively, we can stop the iterations when the relative residual drops below a threshold $\tau$. We define this stopping condition as follows, where $\tau$ is typically on the order of $10^{-2}$ or smaller.

$$(3.40) \qquad \frac{\|f(\mathbf{x}_k) - f(\mathbf{x}^*)\|}{\|f(\mathbf{x}^*)\|} \leq \tau$$

**3.2. The Effect of Noise.** What if the given image $f(\mathbf{x}^*)$ is available not exactly, but rather with uncertainty as $f(\mathbf{x}^*) + $ error? This error may be due to noise, quantization, or some other source of error with bounded magnitude. To simplify matters, let's restrict our attention to the linear filter case where $f(\mathbf{x}) = \mathbf{W}\mathbf{x}$. The intuitions gained from this analysis are applicable to the general (nonlinear) case as well.

First, let's suppose $f(\mathbf{x}^*)$ is given exactly and without error. It is not difficult to show that the rendition algorithm we propose boils down in the linear case to the following simple expression for the $k$-th iterate

$$(3.41) \qquad \mathbf{x}_k = [\alpha \mathbf{I} + \gamma(\mathbf{I} - \mathbf{W})]^k \mathbf{x}_0$$

where $\alpha = 1 - \gamma\mu$, and the intialization is $\mathbf{x}_0 = f(\mathbf{x}^*)$.

Now suppose that we perturb $\mathbf{x}_0$ as $\widetilde{\mathbf{x}}_0 = \mathbf{x}_0 + \Delta\mathbf{x}_0$ where $\Delta\mathbf{x}_0$ is the unknown perturbation. This in turn results in a perturbation of the iterates as follows:

$$(3.42) \qquad \widetilde{\mathbf{x}}_k = [\alpha \mathbf{I} + \gamma(\mathbf{I} - \mathbf{W})]^k \widetilde{\mathbf{x}}_0$$
$$(3.43) \qquad \qquad = \mathbf{x}_k + [\alpha \mathbf{I} + \gamma(\mathbf{I} - \mathbf{W})]^k \Delta\mathbf{x}_0$$

The norm of the error after $k$ iterations is

$$(3.44) \qquad \|\widetilde{\mathbf{x}}_k - \mathbf{x}_k\| = \|[\alpha \mathbf{I} + \gamma(\mathbf{I} - \mathbf{W})]^k \Delta\mathbf{x}_0\|$$
$$(3.45) \qquad \qquad \leq \|[\alpha \mathbf{I} + \gamma(\mathbf{I} - \mathbf{W})]\|^k \|\Delta\mathbf{x}_0\|$$
$$(3.46) \qquad \qquad \leq (\alpha + \gamma(1 + M))^k \|\Delta\mathbf{x}_0\|,$$

where $M$ is the spectral radius[8] of $\mathbf{W}$. Considering the appropriate number of steps

---

[8]Or equivalently the Lipschitz constant of $f(\mathbf{x}) = \mathbf{W}\mathbf{x}$.

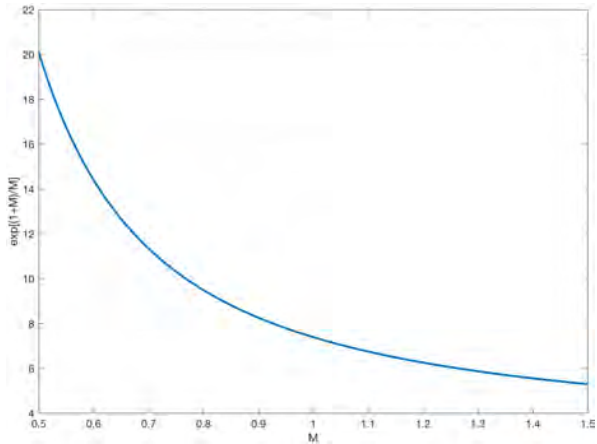

Fig. 3. *Bound on the distortion introduced by unit norm error in measurement of $f(\mathbf{x}^*)$, as a function of the Lipschitz constant $M$.*

to convergence, we have $\gamma = \mathcal{O}(\frac{1}{kM})$ and therefore

$$\|\widetilde{\mathbf{x}}_k - \mathbf{x}_k\| \leq \left(\alpha + \frac{(1+M)}{kM}\right)^k \|\Delta \mathbf{x}_0\| \tag{3.47}$$

$$\leq \left(1 + \frac{(1+M)}{kM}\right)^k \|\Delta \mathbf{x}_0\| \tag{3.48}$$

$$\leq \exp\left(\frac{1+M}{M}\right) \|\Delta \mathbf{x}_0\| \tag{3.49}$$

(3.50)

The exponential factor (see Fig. 3) grows rapidly when $M < 1$ and decays slowly when $M > 1$. This makes intuitive sense: strong blurring operators ($M < 1$) need to accentuate high frequencies in the rendition, and this will amplify the noise. On the other hand, when the distortions are strong sharpening operators ($M > 1$), the rendition will suppress high frequencies, which in turn incidentally limits the effect of noise. In either event, the effect of noise is bounded so long as the perturbation is small and $M$ not far from unity. Near the end of the next section we illustrate some examples.

**4. Experiments.** In this section we present a set of experiments to illustrate and quantify the ability of our proposed method to recover from several forms of degradations including linear and non-linear smoothing, sharpening, the median filter, tone-mapping, and some combinations of these effects.

**4.1. Rendition from Linear and Non-linear Smoothing.** First, let's consider rendition from linear blur. In Figure 4 the image on the left is blurred with a Gaussian point spread function of radius 5 and variance 1. The middle image shows the degraded image. The image on the right hand side (and its corresponding crops shown) are recovered by running the rendition iteration for 35 steps which is when the stopping criterion takes effect. The rendition has improved the PSNR of the degraded image signficantly from 23.82dB to 29.92dB.

The next example illustrated in Figure 5 shows the same image degraded with a disk blur kernel of size $5 \times 5$. The disk PSF has a more severe cutoff, an therefore oscillating frequency response which destroys frequencies that can no longer be recovered. We may therefore expect that the rendition would be less compelling than the Gaussian case as the smoothing effect is more destructive. This is indeed the observation. While the rendition converges in only 26 iterations, the final result has improved the PSNR only by a bit more than 1dB from 22.46 to 23.26dB.

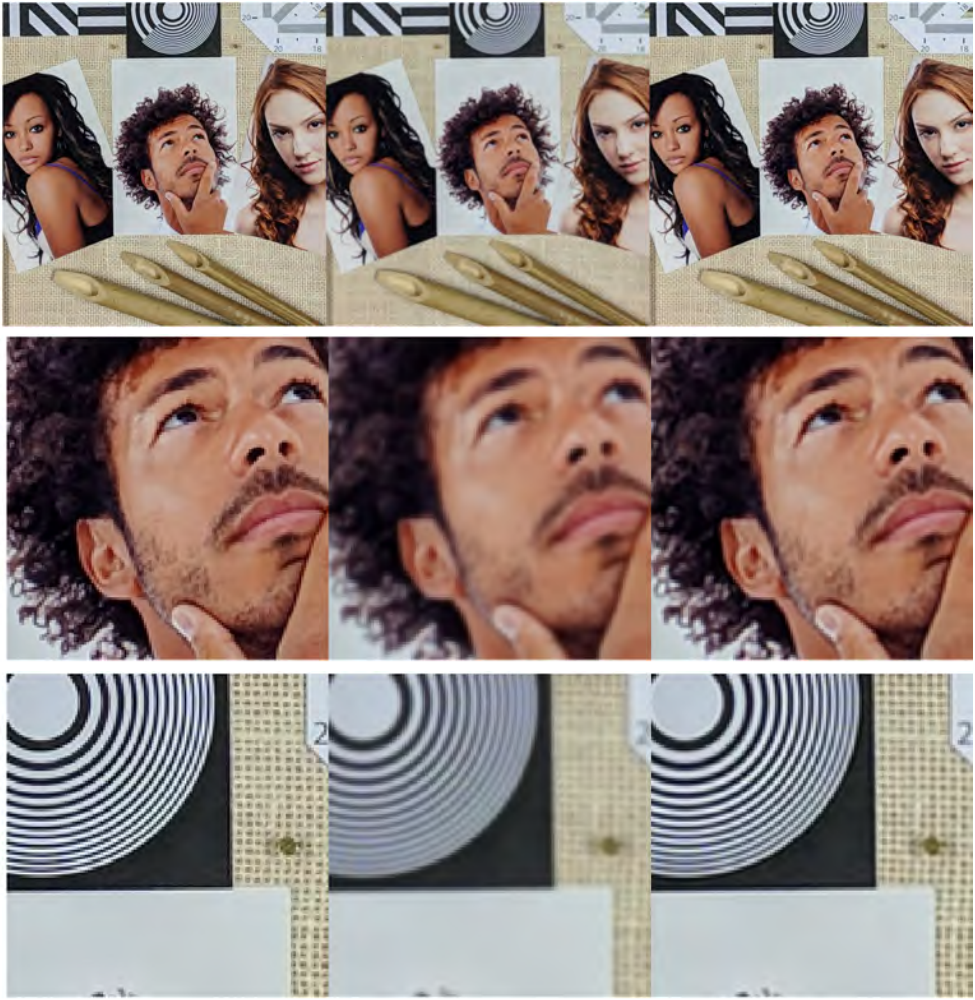

FIG. 4. *Left to right:* $512 \times 512$ *ground truth image; blurred with Gauss(5,1) PSNR = 23.82dB; Rendered PSNR = 30.86dB.* 35 *iterations, step size* $\gamma = 0.15$. *Credit: DxO Labs*

The above experiments illustrate that we can recover from mild (linear) blur of unknown nature rather effectively with the rendition algorithm. The next set of experiments shows what is a more surprising and impressive effect. Namely, we can recover from applying a non-linear filter such as the bilateral, the domain transform, or even the median filter. But the degree of success depends upon the severity of the effect and the type of distortion.

Let's consider the bilateral filter with spatial and range parameters 2 and 1.5

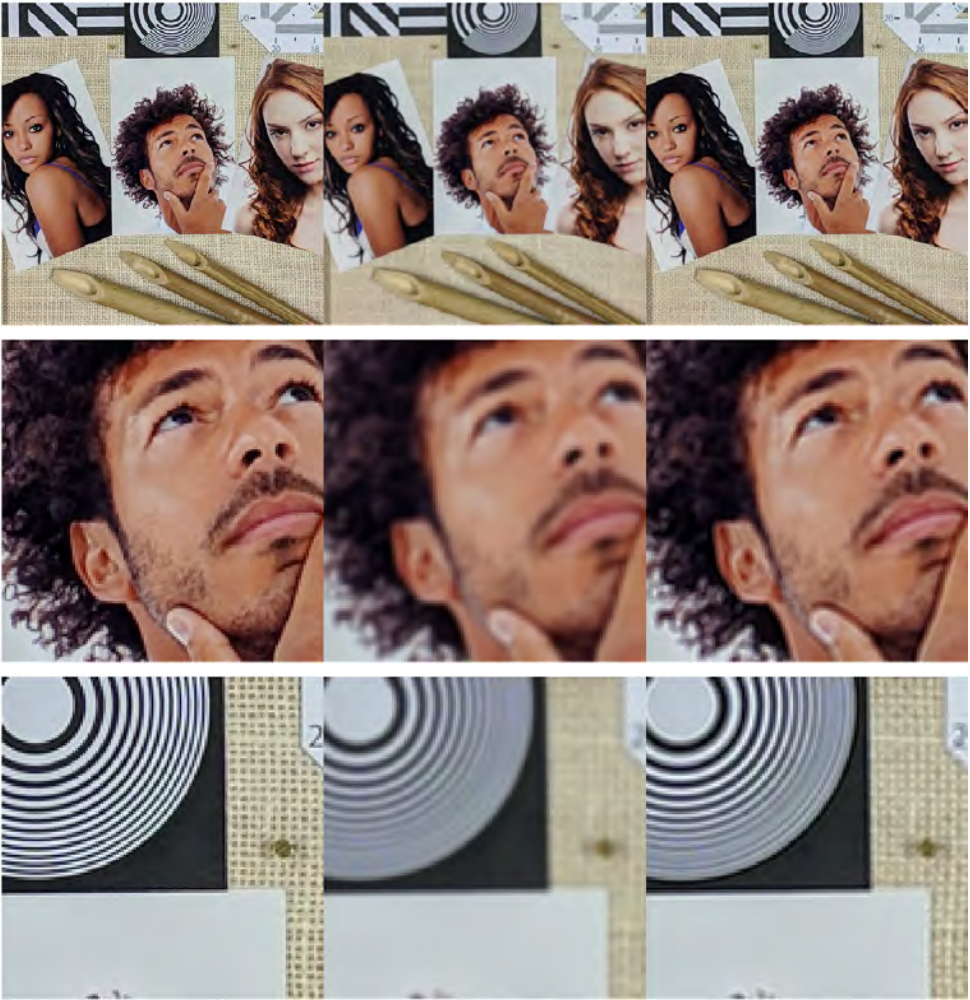

FIG. 5. *Left to right: Left to right: $512 \times 512$ ground truth image; blurred with disk PSF of diameter* 5 *PSNR* $= 22.46 dB$; *Rendered PSNR* $= 23.64 dB$. *Converged after* 26 *iterations, step size* $\gamma = 0.15$

respectively, applied to the image in Figure 6. The more subtle effect of this non-linear smoother is apparent in removing and smoothing some of the smaller scale wrinkles and leaving the larger ones mostly untouched. The rendition after 22 iterations shows many of the small scale features on the face returned to the image. While of coure the rendition is not expected to be perfect, we do observe an impressive increase in PSNR of more than 8dB.

The next example shown in Figure 7 illustrates a median filter of support $2 \times 2$ applied to the image shown on the left. We choose this image in particular for this example since it contains many small scale features that stand out from their surroundings (i.e. apparent outliers). Removing outliers is something the median filter is good at. The smoothing effect is apparent in the middle column. In the right side column, the rendered image and crops can be seen. We note that while the effect

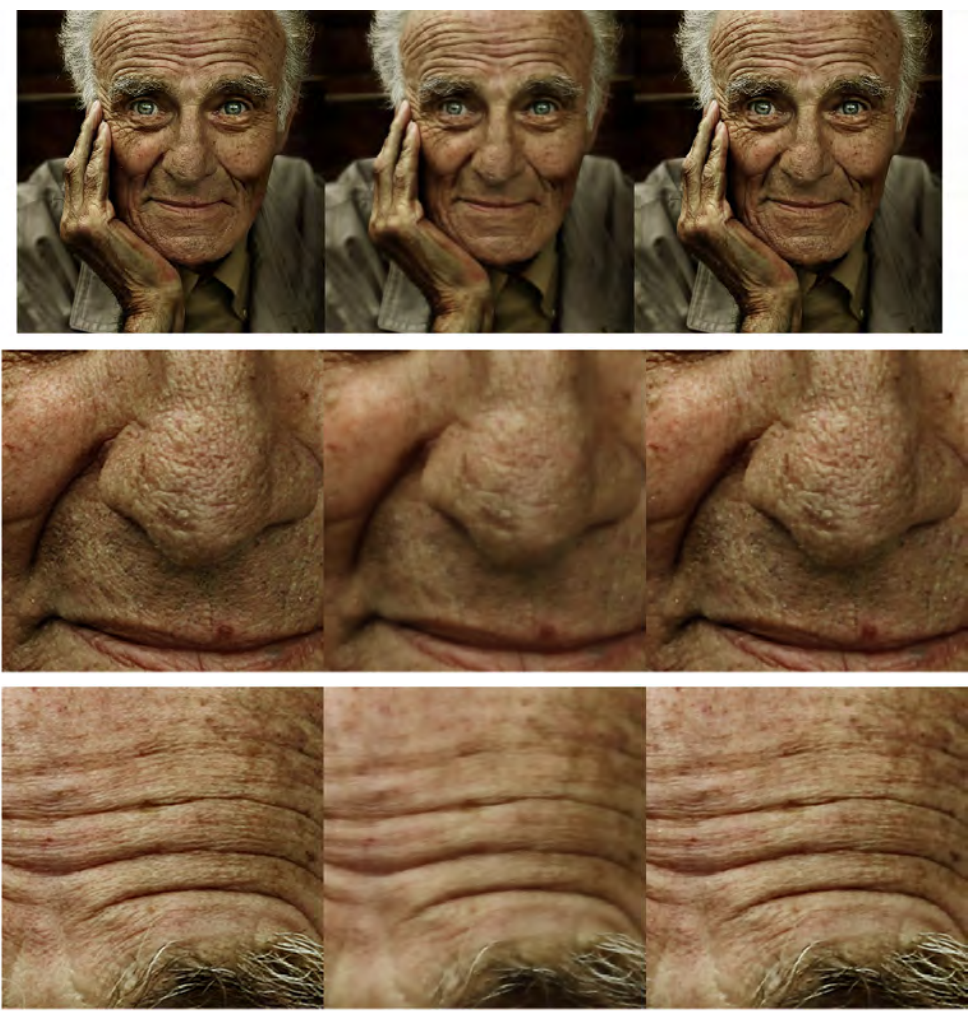

FIG. 6. *Left to right: Ground truth image, Smoothed with Bilateral(**x**,2,1.5) PSNR = 30.00dB; Rendered PSNR = 38.22dB, 22 iterations, step size $\gamma = 0.15$*

is to sharpen the image, the small outliers removed are not effectively returned to the image. Even so, the PSNR has improved by about 3.5dB.

It is reasonable to wonder how far the effects illustrated here can be pushed in their severity and to ask whether we can still get a reasonable rendition. To answer these question, we carry out an experiment shown in Fig. 8 where the strength of the bilateral filter applied to the photo of flowers is increased exponentially in each row. Not only are the parameters of the bilateral filter doubled each time, but the resulting filter is also applied more times in each case. What we observe, not surprisingly, is that the effect of the rendition is less impressive in terms of recovering the original image. Still, it must be said that the effect is not without merit. In fact, it is compelling to see that the rendition does provide sharper images with more well-defined features, even when the filter is quite severe.

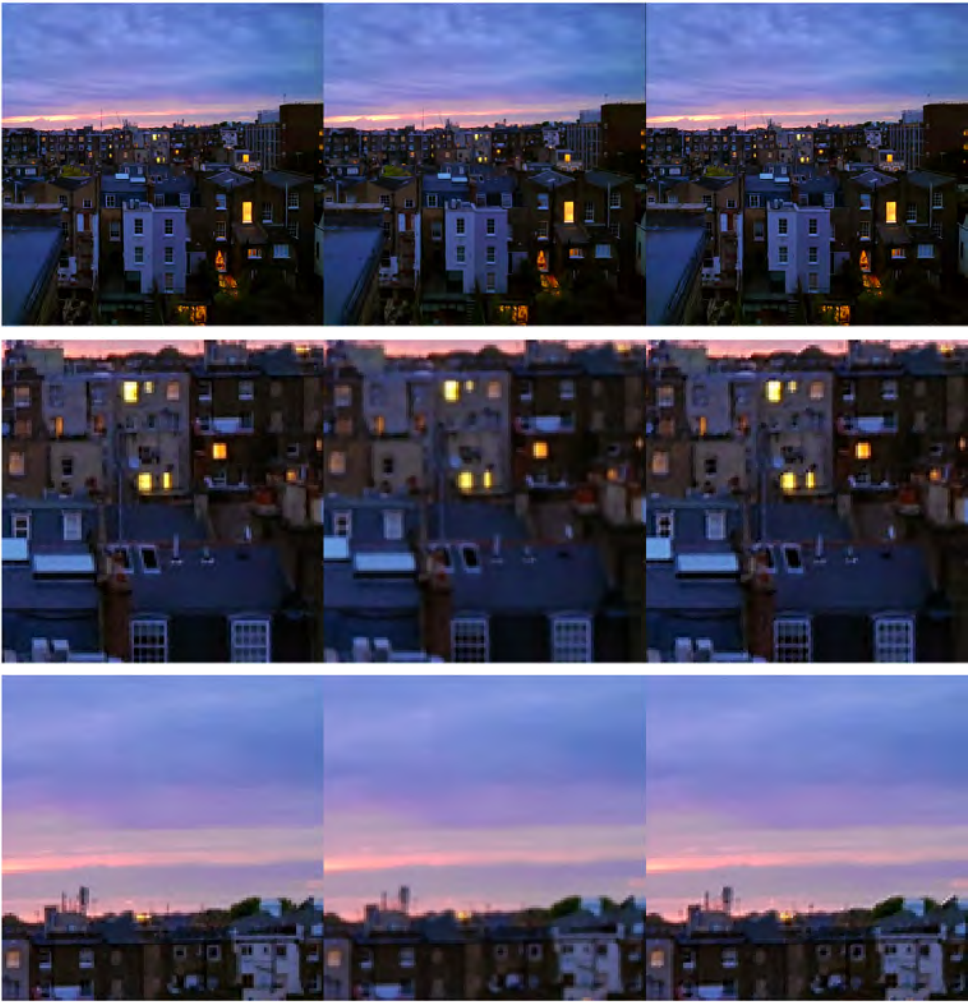

Fig. 7. *Left to right: Ground truth image, Median filtered with Median(**x**,2,2) PSNR 30.52, Rendered PSNR 34.01. 49 iterations, step size $\gamma = 0.04$.* ©: *Peyman Milanfar*

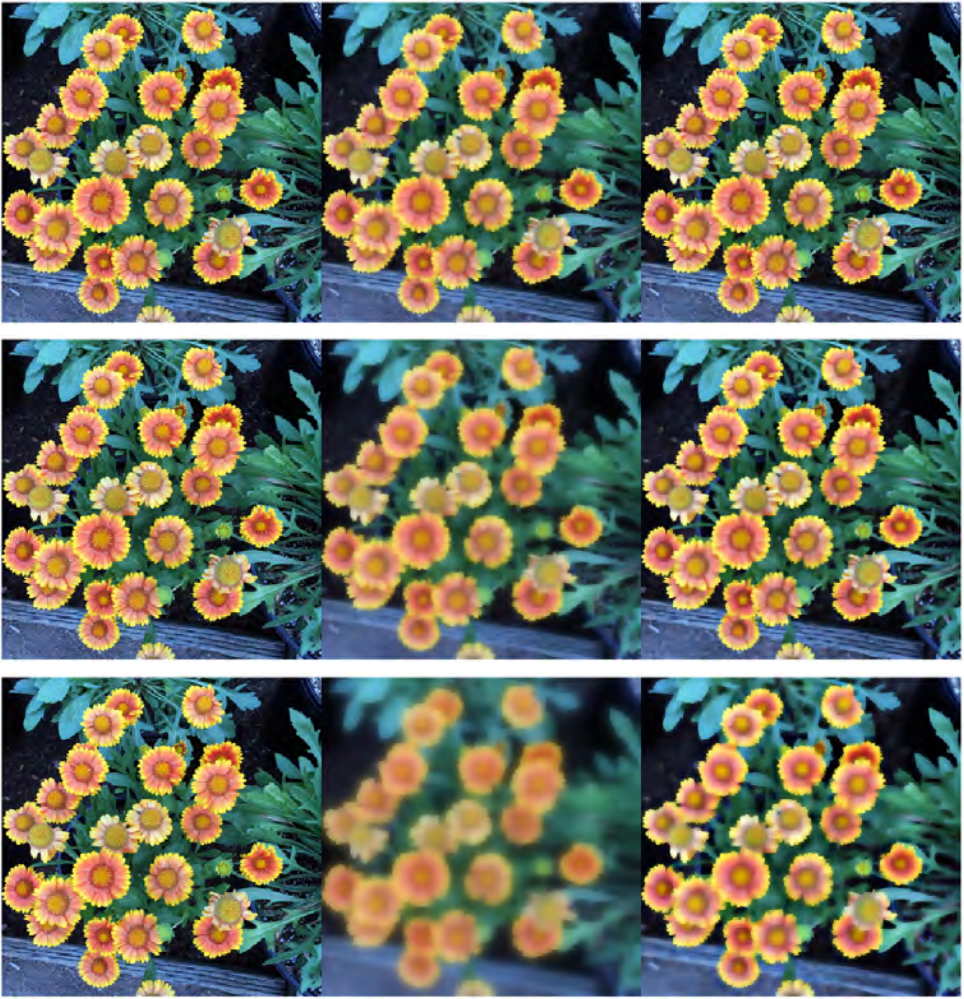

Fig. 8. *Left: Ground truth, Middle: Smoothed with Bilateral Filter. Right: Rendition. From top to bottom: Top: Bilat(2,1) applied 1 time: Redition improves SNR from* 28.74 *to* 37.71 *dB in* 50 *iterations.. Bilat(4,2) applied 2 times: Redition improves SNR from* 22.48 *to* 29.72*dB in* 31 *iterations. Bilat (8,4) applied 4 times: Redition improves SNR from* 18.14 *to* 23.75 *dB in* 50 *iterations. Step size* 0.15*. Credit: Don McCulley, Wikimedia Commons*

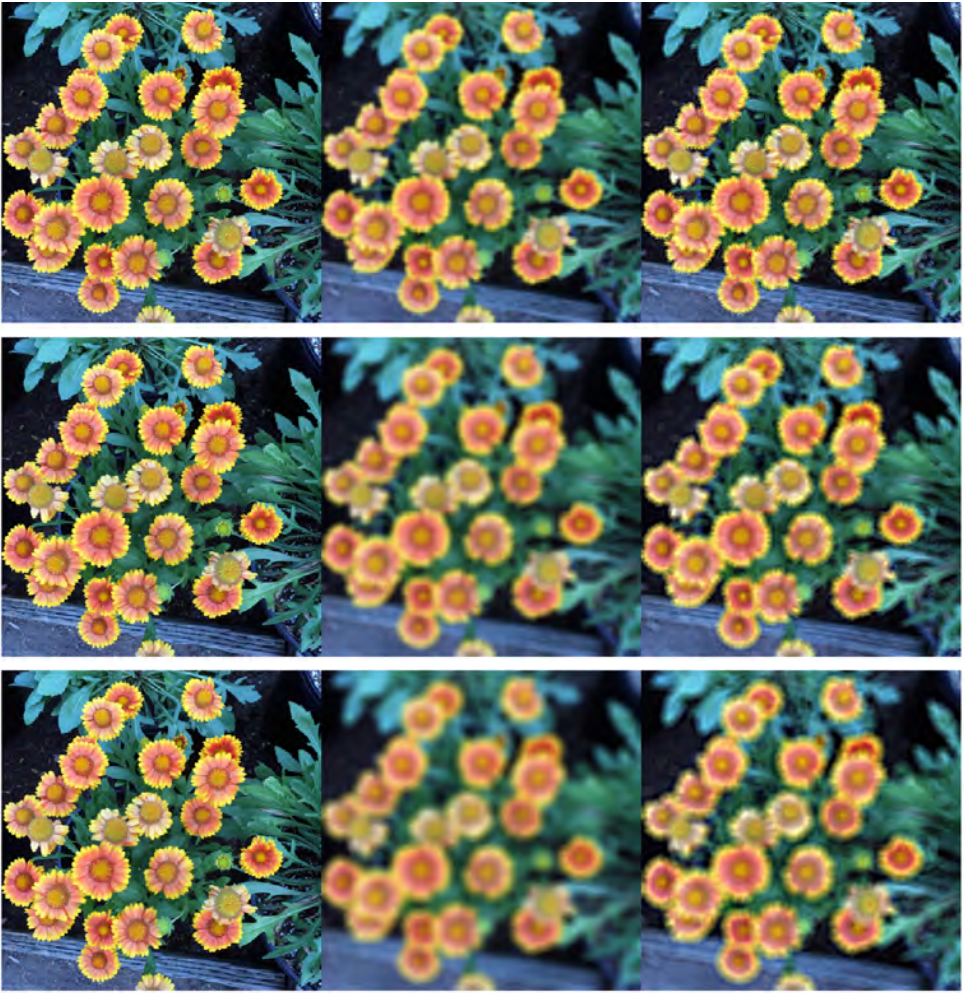

Fig. 9. *Left: Ground truth, Middle: Blurred with Disk PSF of varying sizes. Right: Rendition. Bokeh (Disk) Blur with varying support. From top to bottom: $7 \times 7$: Rendition improves PSNR from $22.19$ to $22.81 dB$ after $13$ iterations; $11 \times 11$: Rendition improves PSNR from $19.8$ to $20.01 dB$ after $23$ iterations; $15 \times 15$ Rendition improves PSNR from $18.43$ to $18.78 dB$ after $23$ iterations. None of these cases improve the PSNR by very much at all, but they all appear to produce a sharper result.*

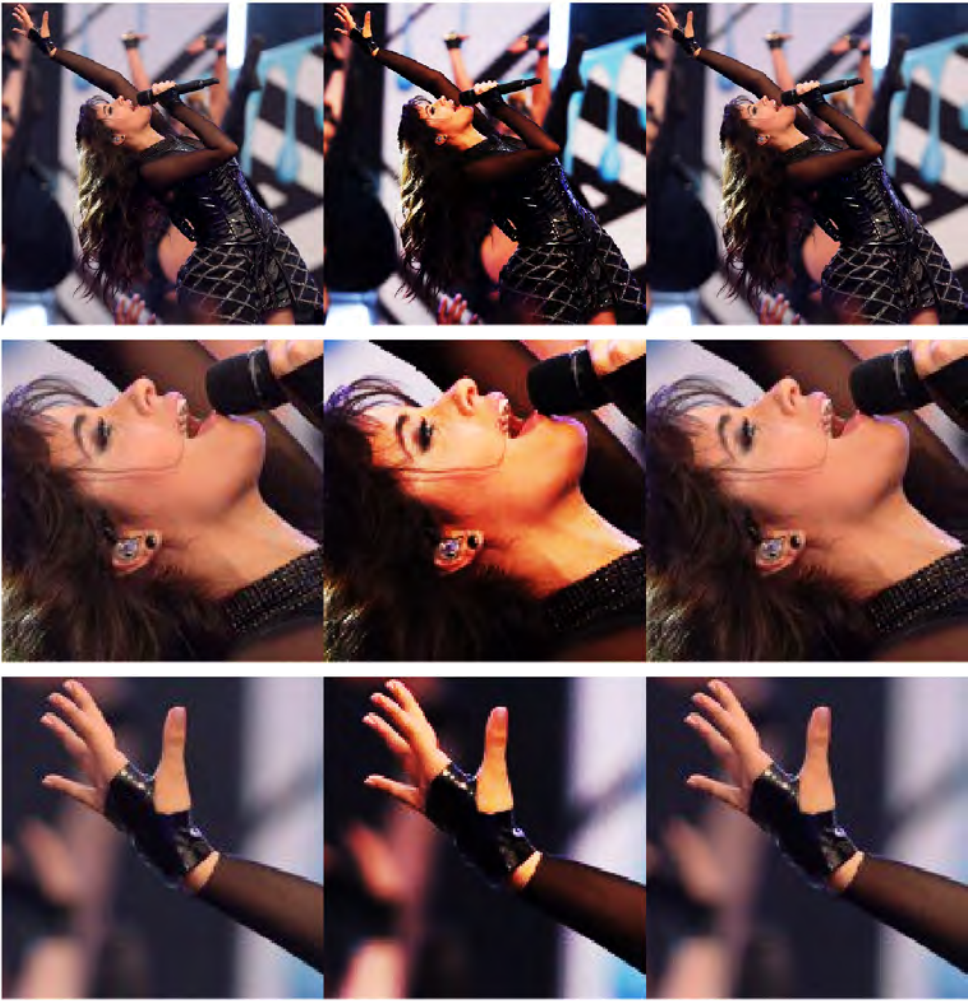

Fig. 10. *Left to right: Ground truth image, Tonemapped in R,G, and B with sigmoid(**x**,0.2) PSNR* 20.77, *Rendered PSNR* 35.87. 20 *iterations, step size* $\gamma = 0.15$. *Credit: Getty Images*

**4.2. Rendition from Tone-mapping, Sharpening, and Composite Effects.** As we advertised earlier in the paper, one of the compelling reasons why rendition is both simple and useful is that it is effective for reversing *non-contractive* effects. Let's consider a common operation that is used to change the look of an image; namely tone-mapping. As illustrated in Fig. 10, the image on the left is passed (in all three color channels) through a nonlinear tone-mapping transformation described in equation (A.1) with parameter $a = 0.2$. The shape of this function can be inferred in Figure 17. The rendition from this tone-mapped image is seen to agree quite well with the ground truth image, as confirmed by the PSNR of 35.87, which is more than 15 dB improvement over the given image.

Next, we consider the case where the image in Fig. 11 is sharpened using a (nonlinear) unsharp mask which employs the bilateral filter. Namely, $f(\mathbf{x}^*) = \mathbf{x}^* + (\mathbf{x}^* -$

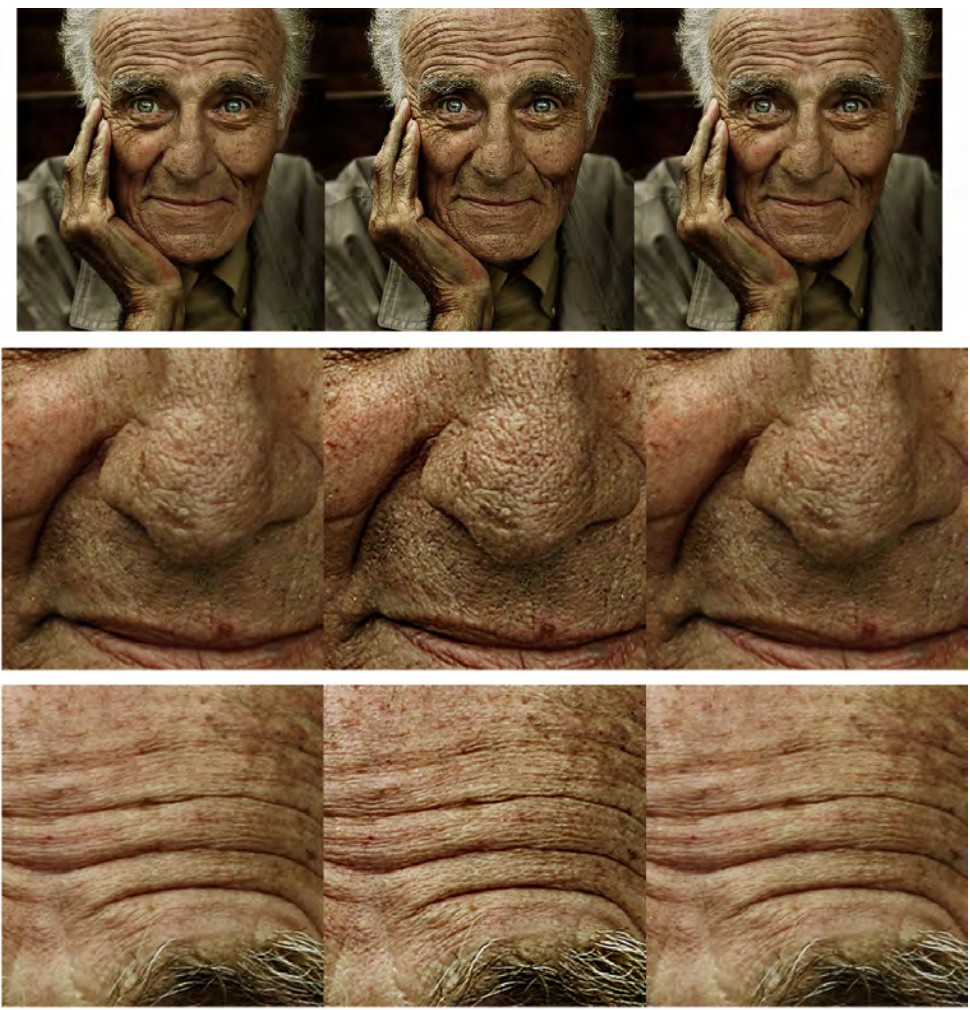

Fig. 11. *Left to right: Ground truth image, Sharpened with 2$\mathbf{x}$ - Bilat($\mathbf{x}$,2,1.5) PSNR = 30.46dB, Rendered PSNR = 45.60dB. 13 iterations, step size $\gamma = 0.15$*

bilat($\mathbf{x}^*, 2, 1.5$)). We seek to render $\mathbf{x}^*$. The ground truth $\mathbf{x}^*$, the given image $f(\mathbf{x}^*)$, and the rendered solution are shown respectively in the left, center, and right panels of Fig. 11. The rendition improves the PSNR by a significant 15 dB margin.

In the next experiment, we employ not only a more severe sharpening, but also apply a strong gamma tone curve to produce the given image as follows: $f(\mathbf{x}^*) = (\mathbf{x}^* + (\mathbf{x}^* - \text{bilat}(\mathbf{x}^*, 10, 3)))^{0.65}$. It is quite encouraging to observe that the rendition is a very close visual approximation of the ground truth image, with a corresponding PSNR of 33.26 dB, having boosted the PSNR by more than 16 dB.

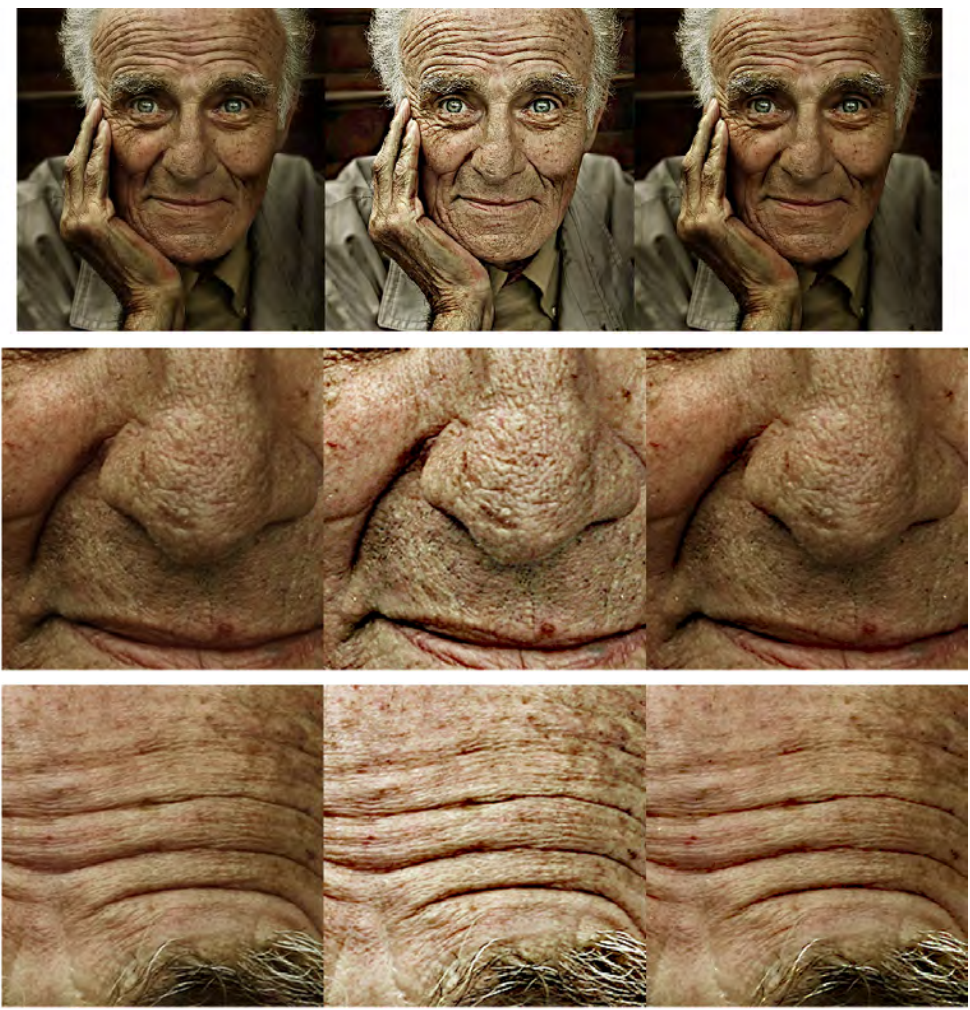

Fig. 12. *Left to right: Ground truth image, Sharpened and gamma corrected with $(2\mathbf{x} - Bilat(\mathbf{x}, 10, 3))^{0.65}$ PSNR 17.19dB. Rendered PSNR 33.26dB. 24 iterations, step size $\gamma = 0.15$*

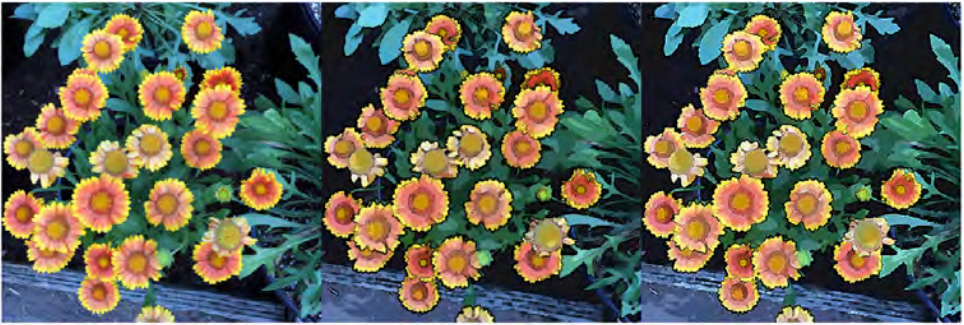

Fig. 13. *An example of an expected failed rendition with $\widehat{M} = 0.67$. Left: Ground truth, Middle: Cartoonized [30]. Right: Best possible rendition shown after 15 iterations. Stopped manually. PSNR improves only from 15.56 to 15.84 dB, and there is no visual improvement. Continuing iterations results in worsening of results – a failed rendition.*

**4.3. Effect of strong perturbations and noise.** As we noted in earlier sections, there are two scenarios in which one can expect the rendition results to either fail completely, or depart visually from the sought-after image. The first occurs if the perturbation is very strong; the second, when the given image is disturbed by noise or error. We examine these scenarios here with a few experiments.

First, let's show a clear example where rendition simply fails to improve the result at all. In Figure 13 we illustrate a cartoonization filter [30] applied to the flower image. This effect is clearly very strong, producing large areas of piecewise constant brightness and color. We would not expect the rendition to be successful. The best possible rendition (in terms of PSNR) is shown after 15 iterations, where the process was terminated manually. The PSNR improves insignificantly from 15.56 to 15.84 dB, and there is no apparent visual improvement. Continuing iterations results in worsening of results – a failed rendition.

Next, we illustrate the effect of noise on the results shown in the previous experiments in this section. As we demonstrated in eq. (3.47), with adequate number of iterations, the effect of perturbation can be bounded, but still significant. In the experiments shown in Fig 14 , we can see how rendition from (Gaussian and Disk) blur are impacted by noise. The variance of the noise here is moderate at 0.05, already reducing the PSNR of the starting image by $2 - 3$ dB. It is interesting to note that in the resulting renditions, the effect of the black box appears to be reversed, but remnants the noise survive in the result.

In Fig. 15, we illustrate the effect of noise on rendition from images either smoothed, or sharpened by the bilateral filter, and in the latter case, also tone mapped. As expected, the effect of noise on the rendition is milder when the black box operator is sharpening, and more noticeable when the operator is purely smoothing.

Finally, in Fig. 16 we show the effect of noise on the rendition from the median filter, direct tone-mapping, and a very strong bilateral filter. All the black box operators here are difficult to invert, and the effect of noise doesn't help matters. In particular, note that in the case of the median filter, the given image is already noisy, and we are studying what happens when we add even more noise. All the above observations are consistent with the bounds in (3.47).

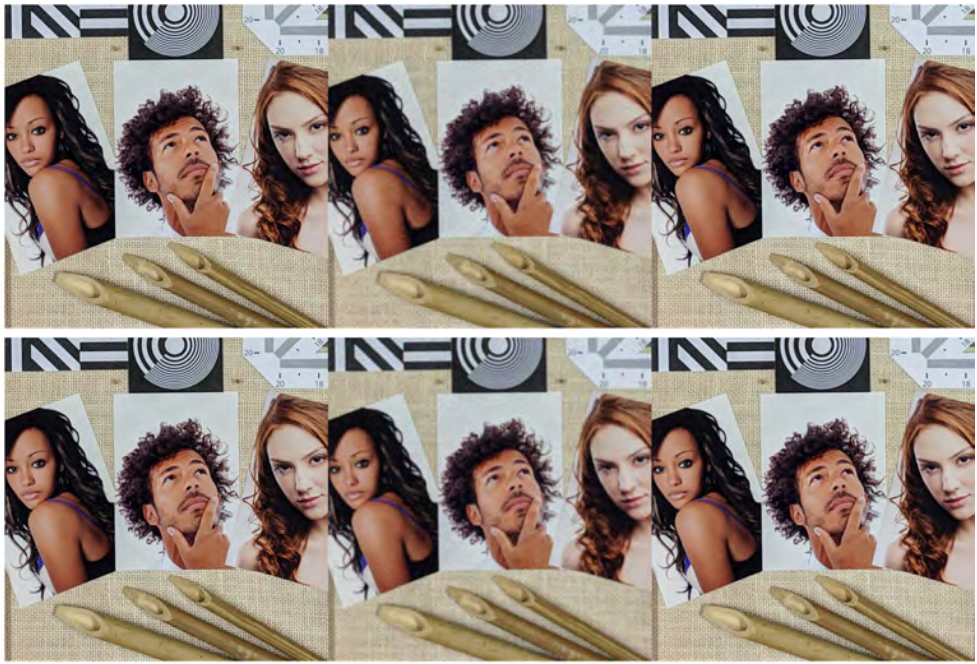

FIG. 14. *Effect of Gaussian white noise with standard deviation* 0.05 *(on a scale of 0 to 1) on rendition from blurring with (first row) Gaussian and (second row) Disk kernels, respectively.*

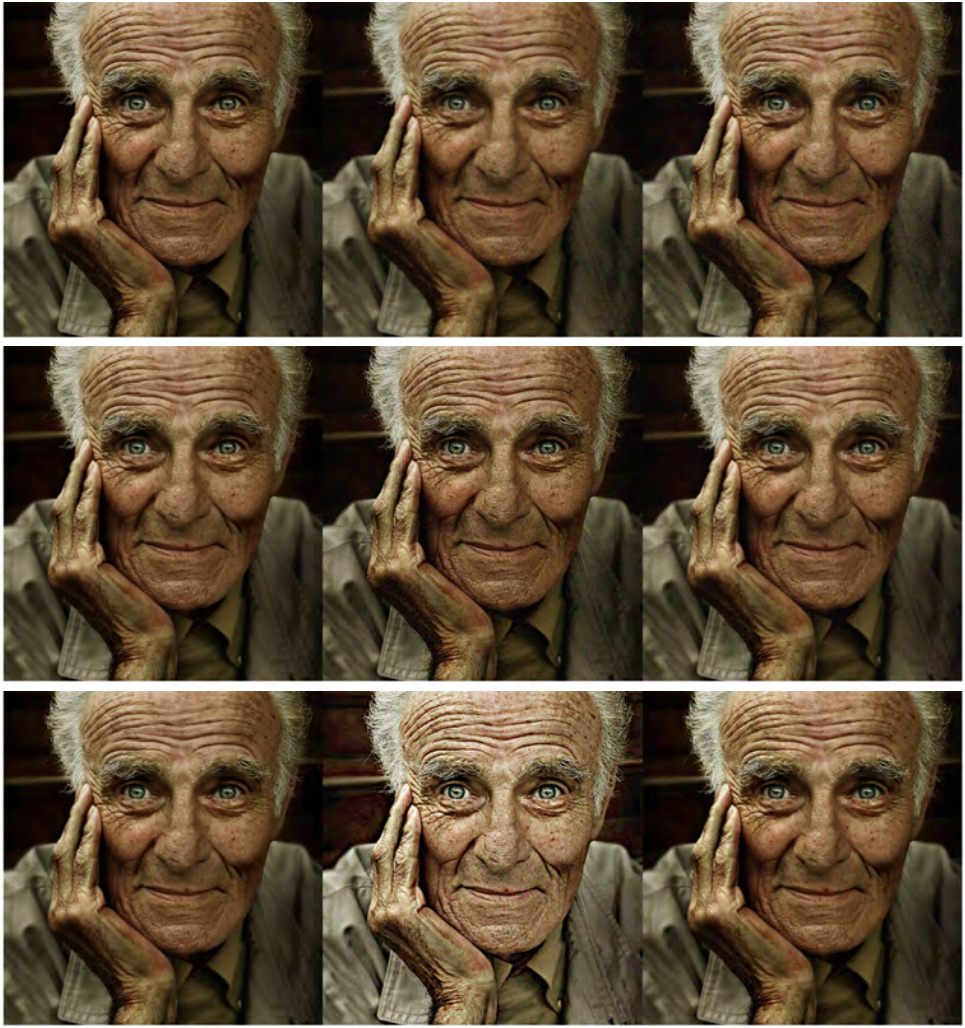

Fig. 15. *Effect of Gaussian white noise with standard deviation* 0.05 *on rendition from bilateral (top row) smoothing, (middle row) sharpening, and (bottom row) combined sharpening and tone-mapping.*

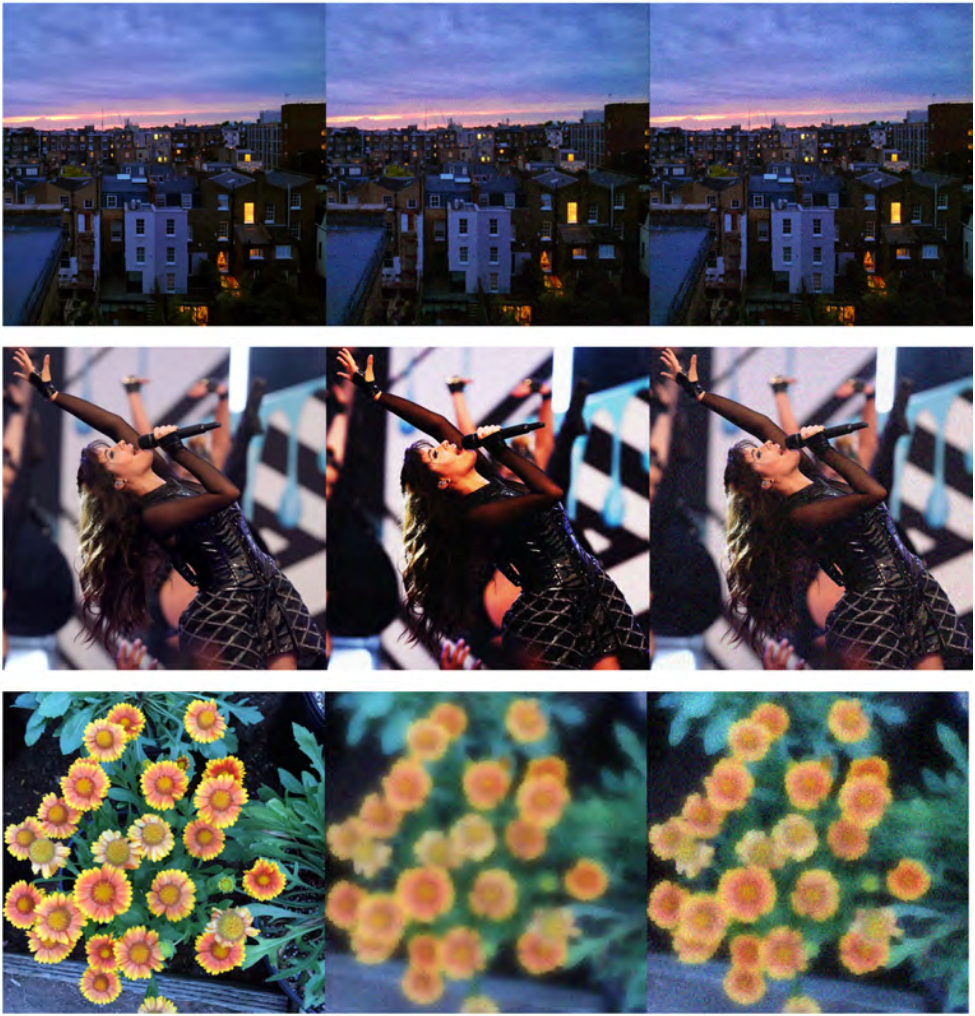

Fig. 16. *Effect of Gaussian white noise with standard deviation* 0.05 *on rendition from (top row) median filter, (middle row) direct tone-mapping, (bottom row) strong bilateral filter.*

**5. Remarks and Conclusions.** In this final section, we summarize again our contributions, and note several areas of further investigation worth considering.

- As we described in the introduction, the premise of the *rendition* problem is very familiar – namely, we wish to invert the effect of an incompletely known process. The problem we've addressed here is quite different from a standard inverse problem in several respects. First, we have not assumed a functional form (e.g. model) for the degradation process. Second, we are not addressing the inversion of a necessarily *physical* forward process. Third, we have not taken advantage of any regularization which would implicitly bring in prior information about the class of images we seek to reconstruct. As such, the description of our problem is rather bare bones and somewhat unorthodox; as is the solution.
- We make weak assumptions (Lipschitz) on the forward operator that distorts the image of interest, and assume access only to *evaluation* of this operator, not its "insides". We proved that as long as the damage done by this operator to the unobserved input is not too heavy, we can roughly undo its effect and recover a good rendition of the input. Notably, we showed that the resulting algorithm becomes very simple, resembling a fixed point operation. However, this algorithm is shown to be convergent for a far wider class of perturbations than simply "contractive" ones. Indeed, a very large variety of practical operations such as nonlinear local sharpening and tone-mapping which are applied by image processing pipelines are non-contractive, yet apparently reversible so long as these effects are not too strong.
- A final question worth asking is whether standard approaches to solving inverse problems such as regularization are still relevant to rendition. Conceptually, the answer is yes; but it is less clear how effective regularization would be. Consider a regularization term added to our rendition loss function:

$$\text{(5.1)} \quad J_\lambda(\mathbf{x}) = \phi(\mathbf{x}) + \lambda \, \rho(\mathbf{x})$$
$$\text{(5.2)} \quad = \mathbf{x}^T(f(\mathbf{x}) - f(\mathbf{x}^*)) - \frac{1}{2}\mathbf{x}^T\mathbf{x} + \lambda \, \rho(\mathbf{x})$$

The regularization term with strength $\lambda$ can take any number of forms as usual. A particularly straightforward case to analyze arises if we use the framework of regularization by denoising (RED) [23]. This form of regularization can take advantage of any denoising (smoothing, non-expansive) function $s(\mathbf{x})$. The RED regularizer (which is *convex*), and its gradient [23] are

$$\text{(5.3)} \quad \rho(\mathbf{x}) = \frac{1}{2}\,\mathbf{x}^T(\mathbf{x} - s(\mathbf{x})),$$
$$\text{(5.4)} \quad \nabla\rho(\mathbf{x}) = \mathbf{x} - s(\mathbf{x}).$$

Therefore, the gradient of the regularized loss $J_\lambda(\mathbf{x})$ is

$$\text{(5.5)} \quad \nabla J_\lambda(\mathbf{x}) = \nabla\phi(\mathbf{x}) + \lambda \, \nabla\rho(\mathbf{x})$$
$$\text{(5.6)} \quad = f(\mathbf{x}) - f(\mathbf{x}^*) + \mu \, \mathbf{x} + \lambda(\mathbf{x} - s(\mathbf{x}))$$

And hence, the gradient descent applied to the regularized loss is given by

$$\text{(5.7)} \quad \mathbf{x}_{k+1} = \mathbf{x}_k - \gamma\left[f(\mathbf{x}_k) - f(\mathbf{x}^*) + \mu\mathbf{x} + \lambda(\mathbf{x}_k - s(\mathbf{x}_k))\right]$$

We can rewrite this algorithm in a way that is more illuminating:

$$(5.8) \qquad \mathbf{x}_{k+1} = (1 - \gamma(\mu + \lambda)) \mathbf{x}_k + \gamma\lambda \, s(\mathbf{x}_k) - \gamma \left[ f(\mathbf{x}_k) - f(\mathbf{x}^*) \right]$$

In comparison to the un-regularized solution ($\lambda = 0$) originally shown in (3.31), we note that the dampening in the first term ($\mathbf{x}_k$) on the right-hand-side has been strengthened, and a second smoothing term $\gamma\lambda s(\mathbf{x}_k)$ has been added. The residual term (which does most of the work in rendition) meanwhile remains unaffected. Therefore, while the iterative procedure is expected to me more robust, we would not expect this change to have a big impact on improving the visual quality of the rendition. Our preliminary experiments confirm this observation. Further analysis of convergence, along with the use of different regularization functions may still be worth investigating.

**Appendix A. Example computations of the Lipschitz constant.**
- *Linear Filtering:*
    - Let $f(\mathbf{x}) = \text{blur}(\mathbf{x}, \mathbf{h})$ where $\mathbf{h}$ is a point-spread function. Using[9] $\mathbf{h} = \text{fspecial('gaussian',2,1)}$, we have a 2×2 box filter which yields $\widehat{M} = 0.867$. Increasing the blur strength to $\mathbf{h} = \text{fspecial('gaussian',5,3)}$, we get a smaller $\widehat{M} = 0.825$. Increasing further still $\mathbf{h} = \text{fspecial('gaussian',15,5)}$, we get $\widehat{M} = 0.818$
    - Now consider an unsharp masking operation as $f(\mathbf{x}) = \mathbf{x} + \alpha(\mathbf{x} - \text{blur}(\mathbf{x}, \mathbf{h}))$ where $\mathbf{h} = \text{fspecial('gaussian',5,3)}$ and $\alpha = 0.5$. This gives $\widehat{M} = 1.05$. Increasing the blur strength to $\mathbf{h} = \text{fspecial('gaussian',15,5)}$ and the blending strength to $\alpha = 1.0$ gives $\widehat{M} = 1.078$
    - Next, let's consider downscaling and then upscaling a given image by some integer factor $q$. Namely, $f(\mathbf{x}) = \text{resize}(\text{resize}(\mathbf{x}, 1/q), q)$. Suppose we use bi-cubic resampling and $q = 2$. This gives $\widehat{M} = 0.841$. With Lanczos resampling the result is a bit sharper so the estimate the Lipschitz constant is a bit higher $\widehat{M} = 0.850$. Keeping the Lanczos resampler, but increasing $q = 4$, we expect the process to lose more high frequencies and therefore result in a smaller $\widehat{M} = 0.826$
- *Nonlinear Filtering:*
    - Consider the domain transform [12] (which is an approximation of the bilateral filter) $f(\mathbf{x}) = \text{bilat}(\mathbf{x}, \sigma_1, \sigma_2)$, where $\sigma_1, \sigma_2$ are the spatial and range smoothing parameters. Letting $[\sigma_1, \sigma_2] = [2, 0.5]$, we estimate $\widehat{M} = 0.981$. Increasing the smoothing level to $[\sigma_1, \sigma_2] = [5, 1]$, we estimate $\widehat{M} = 0.886$.
      Now let's construct *nonlinear* unsharp masks using this operator. Analogous to the earlier linear case $f(\mathbf{x}) = \mathbf{x} + \alpha(\mathbf{x} - \text{bilat}(\mathbf{x}, \sigma_1, \sigma_2))$. Letting $[\sigma_1, \sigma_2] = [5, 1]$, and $\alpha = 0.4$, we get $\widehat{M} = 1.02$; where as $[\sigma_1, \sigma_2] = [10, 3]$, $\alpha = 1$ gives $\widehat{M} = 1.083$.
    - Next consider the median filter $f(\mathbf{x}) = \text{median}(\mathbf{x}, \mathbf{w})$ where $\mathbf{w}$ denotes the dimensions of the filter footprint. Letting $\mathbf{w} = [3, 3]$ gives a $3 \times 3$ median filter and $\widehat{M} = 1.11$. It is interesting to note that since the median filter tends to produce very sharp edges, it is *not* a contractive filter even for small window sizes. A larger window size $\mathbf{w} = [7, 7]$ gives an even larger $\widehat{M} = 1.162$. The relatively large values of $M$ estimated

---

[9]Matlab notation

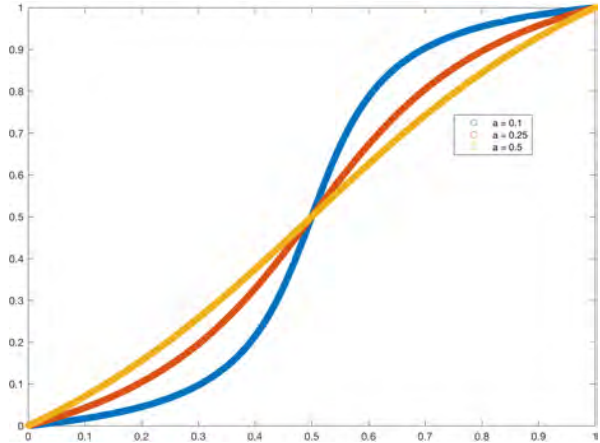

Fig. 17. *Examples of tone-mapping functions*

for the median filter indicate that this is a rather difficult effect to undo. We will verify this in the experimental section.

- *Tone-mapping and Gamma Correction:* We often use these pointwise nonlinear operators $f(x_i) = T(x_i)$ to contract or expand the dynamic range of the image, while remaining still in the range $[0, 1]$, or to do histogram equalization or modification.
    - Let's consider the sigmoid-type operator that rescales the brightness levels of an image, but maps 0.5 to itself:

$$(A.1) \qquad f(x_i) = \sigma(x_i, a) = \frac{\mathrm{atan}(1/2a) + \mathrm{atan}((x_i - 0.5)/a)}{2\mathrm{atan}(1/2a)}$$

    With $a = 0.1$ we estimate $\widehat{M} = 1.239$, whereas with $a = 0.25$ we get $\widehat{M} = 1.12$, and with $a = 0.5$ we get $\widehat{M} = 1.049$. The respective tone-mapping curves are shown in Fig. 17.

- *Compression:* Here we consider applying jpg compression to an image. Namely $f(\mathbf{x}) = \mathrm{jpeg}(\mathbf{x}, q)$ where $1 \le q \le 100$ is the quality factor. We have

$$\widehat{M}([10{:}10{:}100]) = [0.891, 0.911, 0.924, 0.923, 0.92, 0.92, 0.916, 0.91, 0.904, 0.901, 0.90].$$

We observe that interestingly, while jpeg compression is always a contractive operator, the Lipschitz constant may not be a monotonic function of the quality factor.